\documentclass{article}
\usepackage{spconf,amsmath,graphicx,hyperref}
\usepackage{multirow}
\usepackage{xcolor}
\usepackage{enumitem}
\usepackage[utf8]{inputenc}

\let\titleold\title
\renewcommand{\title}[1]{\titleold{#1}\newcommand{\thetitle}{#1}}
\def\maketitlesupplementary
   {
   \newpage
       \twocolumn[
        \centering
        \Large
        \textbf{\thetitle}\\
        \vspace{0.5em}Supplementary Material \\
        \vspace{1.0em}
       ] 
   }

\title{Losing the Plot: How VLM responses degrade on imperfect charts}
\name{\parbox{\textwidth}{\centering Philip Wootaek Shin\textsuperscript{1,2}, Jack Sampson\textsuperscript{1}, Vijaykrishnan Narayanan\textsuperscript{1},\\ 
\textit{Andres Marquez}\textsuperscript{2}, \textit{Mahantesh Halappanavar}\textsuperscript{2}}}

\address{\textsuperscript{1}The Pennsylvania State University, \textsuperscript{2}Pacific Northwest National Laboratory}

\begin{document}
\ninept
\maketitle

\begin{abstract}

Vision–language models (VLMs) show strong results on chart understanding, yet existing benchmarks assume clean figures and fact-based queries. Real-world charts often contain distortions and demand reasoning beyond simple matching. We evaluate ChatGPT-4o, Claude Sonnet 4, and Gemini 2.5 Pro, finding sharp performance drops under corruption or occlusion, with hallucinations such as value fabrication, trend misinterpretation, and entity confusion becoming more frequent. Models remain overconfident in degraded settings, generating plausible but unsupported explanations.

To address this gap, we introduce CHART NOISe(\textbf{C}hart \textbf{H}allucinations, \textbf{A}nswers, and \textbf{R}easoning \textbf{T}esting on \textbf{N}oisy and \textbf{O}ccluded \textbf{I}nput \textbf{Se}lections), a dataset combining chart corruptions, occlusions, and exam-style multiple-choice questions inspired by Korea’s CSAT English section. A key innovation is prompt reverse inconsistency, where models contradict themselves when asked to confirm versus deny the same statement. Our contributions are threefold: (1) benchmarking state-of-the-art VLMs, exposing systematic vulnerabilities in chart reasoning; (2) releasing CHART NOISe, the first dataset unifying corruption, occlusion, and reverse inconsistency; and (3) proposing baseline mitigation strategies such as quality filtering and occlusion detection. Together, these efforts establish a rigorous testbed for advancing robustness and reliability in chart understanding.

\end{abstract}
\begin{keywords}
Chart Understanding, Vision–Language Models (VLMs), Robustness Evaluation, Prompt Reverse Inconsistency, Corruption and Occlusion
\end{keywords}
\section{Introduction}
\label{sec:intro}

Chart understanding remains a critical, yet underexplored, challenge for vision–language models (VLMs). Existing benchmarks often evaluate models on clean, well-structured figures\cite{ChartQARL,wang2024charxiv}, but they fail to capture the complexity of real-world settings where charts are corrupted, partially occluded, or paired with reasoning-intensive questions. 

To address this challenge, we introduce a new evaluation framework that systematically probes VLM performance under chart corruptions, occlusions, and reasoning-intensive question formats. Unlike prior chart benchmarks that emphasize surface-level question answering, our design incorporates multiple choice exam-style questions inspired by Korea’s CSAT English section\cite{kice}, which explicitly requires identifying incorrect statements about charts. This setting forces models to go beyond simple fact-matching and instead demonstrate precise reasoning and interpretation, even under visual distortions.

\begin{figure*}[t]
    \begin{center}
        \includegraphics[width=0.85\linewidth]{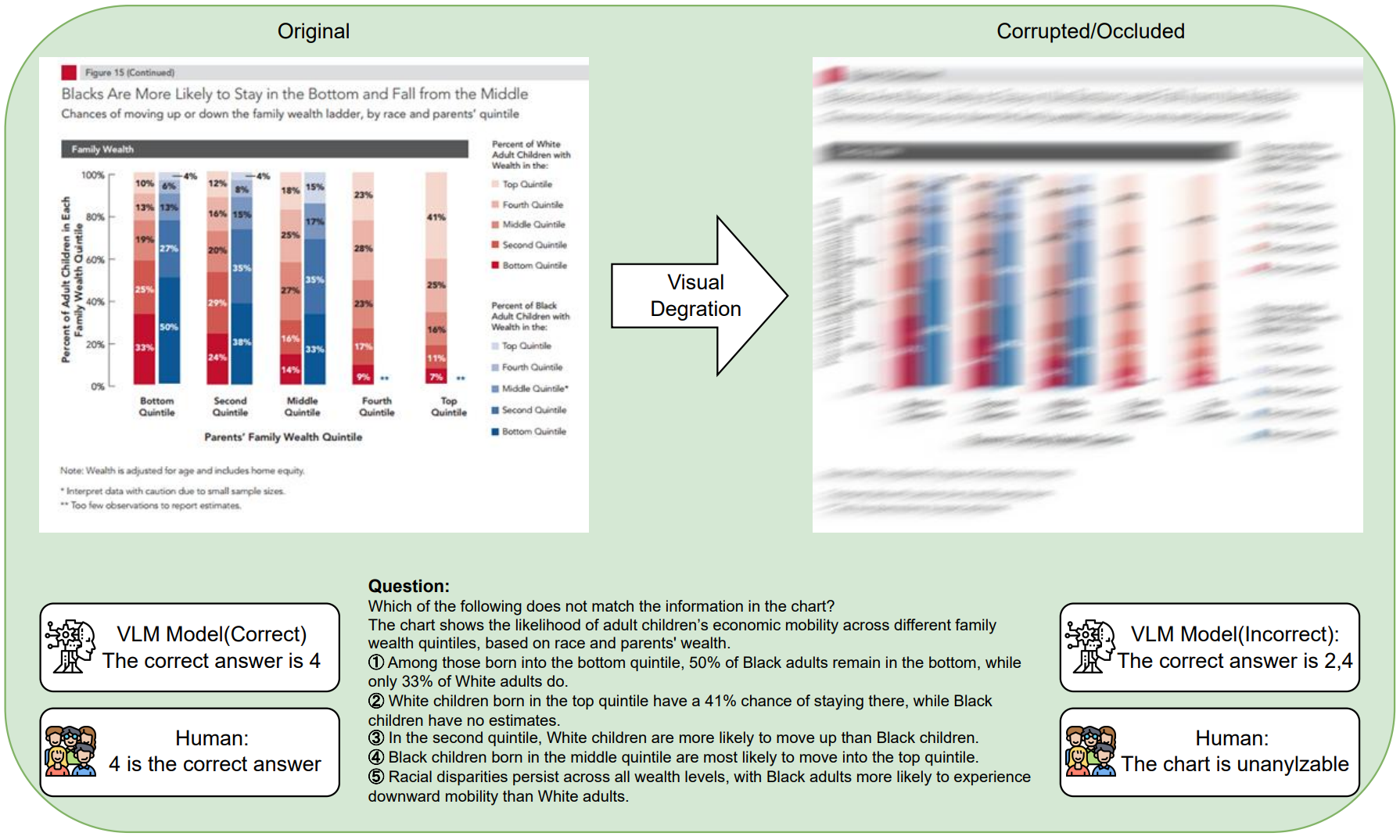}
    \end{center}
\caption{ Example from the CHART NOISe dataset: although the corrupted chart is difficult for humans to interpret, the VLM still attempts to answer.}
\label{fig:ProblemStatement}
\end{figure*}

Fig.~\ref{fig:ProblemStatement} illustrates the core idea of our dataset: integrating chart corruptions and occlusions with exam-style multiple-choice questions that require identifying incorrect statements\cite{Reverseprompt}. This design exposes how VLMs handle both visual degradation and reasoning demands, providing a more realistic and rigorous evaluation of chart comprehension.

Our contributions are threefold: (1) we benchmark state-of-the-art VLMs\cite{gemini2.5pro,chatgpt4o,claude-sonnet4}, revealing systematic hallucination patterns such as value fabrication and trend misinterpretation under degraded conditions; (2) we introduce CHART NOISe (\textbf{C}hart \textbf{H}allucinations, \textbf{A}nswers, and \textbf{R}easoning \textbf{T}esting on \textbf{N}oisy and \textbf{O}ccluded \textbf{I}nput \textbf{Se}lections), a dataset of CSAT-style chart questions paired with diverse corruptions and occlusions (1,500 charts), enabling controlled robustness analysis; and (3) we propose baseline strategies for mitigating these issues, including front-end quality filtering and occlusion detection. Together, these efforts establish CHART NOISe as the first rigorous testbed for evaluating chart reasoning robustness in real-world settings.

\section{Background}
\label{sec:background}

\subsection{Chart specific models}
\label{subsec:Chart}

A complementary line of work focuses on translating charts into structured tabular or textual formats before performing reasoning. DePlot\cite{deplot} decomposes chart reasoning into two steps: (i) plot-to-text translation, and (ii) reasoning over the translated representation, using a modality conversion module as the core component. UniChart\cite{unichart} introduces chart-specific pretraining tasks, combining low-level extraction of visual elements (e.g., bars, lines, data points) with high-level chart understanding and reasoning objectives. ChartAssistant\cite{chartassisstant} employs a two-stage pipeline, consisting of chart-to-table parsing pretraining for modality alignment, followed by multitask instruction-tuning for reasoning. More recently, TinyChart\cite{tinychart} demonstrated that efficient architectures can outperform larger 13B MLLMs on multiple benchmarks, achieving state-of-the-art results with substantially faster inference.

Despite their progress, these approaches share notable limitations. By relying on chart-to-table translation as an intermediate step, they risk error propagation from imperfect parsing, particularly when textual elements or fine-grained visual details are distorted. Moreover, translation-based methods often oversimplify reasoning by converting visual tasks into purely textual ones, overlooking robustness to visual corruptions, occlusions, or layout variability—factors critical in real-world chart comprehension.



\subsection{Vision language model hallucination}

Hallucination in Vision–Language Models (VLMs) has been widely surveyed \cite{survey1,survey2}, with several approaches proposed to mitigate its effects \cite{park2025second,zhao2024mitigating,liu2024reducing}. Recent efforts include pioneering frameworks such as BaviBench \cite{BaviBench} and versatile evaluation tools that probe robustness by introducing adversarial instructions.

For chart data, hallucination can take multiple forms, including but not limited to: value fabrication (producing incorrect numerical values), trend misinterpretation (misreading upward or downward patterns), entity confusion (misidentifying chart elements), reasoning hallucination (introducing unsupported logical steps), and table/translation drift (errors during chart-to-table conversion). These errors are further amplified under visual corruptions such as noise or occlusion.

Given the significance of hallucination in chart understanding, our work introduces a dataset designed to systematically evaluate VLM robustness to both chart corruptions and complex hallucination scenarios.


\subsection{Dataset}
Several datasets evaluate chart understanding in vision–language models (VLMs). PlotQA\cite{methani2020plotqa} is the largest, with 28.9M question–answer pairs over 224,377 plots, but its reliance on templated questions limits reasoning diversity. ChartQAPro\cite{ChartQAPro} offers 1,341 charts from 157 sources with multiple reasoning modes (conversational, multiple-choice, hypothetical, fact-checking), though its scale and chart complexity remain modest. ChartQA-RL\cite{ChartQARL} augments the original ChartQA\cite{masry2022chartqa} by removing textual labels and adding perturbations, improving visual robustness evaluation but reducing text–visual reasoning opportunities. ChartXiv\cite{wang2024charxiv} contributes 2,323 naturally occurring charts from arXiv papers, enhancing realism but limited in scale and alignment with standardized testing.

Overall, current datasets vary in size, realism, and reasoning scope, yet none address both robustness to noise/occlusion and complex reasoning comparable to standardized exams. This motivates our dataset, which integrates corruptions, occlusions, and challenging reasoning tasks to rigorously assess VLM performance.



\section{Dataset Generation}
\label{sec:benchmark}

We were motivated by the phenomenon of prompt reverse inconsistency\cite{Reverseprompt}, where LLMs often yield conflicting answers when asked “Which are correct?” versus “Which are incorrect?”. This inspired us to design chart-based evaluation beyond simple chart-matching tasks commonly found in existing datasets. Notably, Korea’s CSAT English exam\cite{kice} includes at least one question requiring identification of an incorrect statement about a chart, demanding interpretation and reasoning rather than surface-level matching. However, the charts in these exams are relatively simple compared to the complexity of the accompanying questions. Building on this, we sought to create more challenging questions that combine prompt reverse inconsistency with complex charts and tables, enabling a deeper evaluation of state-of-the-art VLM performance.

\subsection{Baseline Test}
\label{subsec:Baseline}
We collected chart data from the ChartQAPro\cite{ChartQAPro} dataset by randomly sampling 200 charts and selecting 100 that were sufficiently complex to support question generation. For each chart, we used ChatGPT and Gemini APIs to generate Korea CSAT–style multiple-choice questions. Questions deemed too simple and incorrect were revised by the authors, resulting in a total of 100 validated questions across 100 charts. For comparison, we incorporated chart-based questions  from the English section of Korean high school examinations. For grade 1 and grade 2, four exams are administered annually by the Office of Education of four different districts (March–Seoul, June–Busan, September–Incheon, November–Gyeonggi-do) from March 2013 to June 2025, yielding 50 pairs each. For grade 3, while June, September, and November exams are administered by KICE\cite{kice}, four sessions are conducted by the Office of Education of four different districts (March–Seoul, April–Gyeonggi-do, July–Incheon, October–Seoul) from March 2010 to June 2025, producing 108 pairs

\begin{table}[t]
    \centering
    \footnotesize
    \resizebox{\columnwidth}{!}
    {   
    \begin{tabular}{l|ccc}
        \hline
        & 
        \textbf{Chatgpt4o} &  \textbf{Gemini2.5Pro} &\textbf{ClaudeSonnet4} \\
        \hline
        CSAT12 & 97/100 & 100/100 & 97/100  \\
        CSAT3 & 95/108  & 102/108 & 98/108 \\
        \textbf{CHART NOISe(Orig)} & 70/100  & 88/100 & 76/100 \\
        \hline
    \end{tabular}
    }
\caption{Comparison of model performance on CSAT benchmarks and our CHART NOISe dataset without transformations(Orignial), measured by accuracy}
\label{CSATtable}

\end{table}

We evaluated the generated questions using three models (ChatGPT-4o\cite{chatgpt4o}, Gemini 2.5 Pro\cite{gemini2.5pro}, and Claude Sonnet 4\cite{claude-sonnet4}) across Grades 1, 2, and 3. Grades 1 and 2 were combined, as their charts are relatively simple compared to Grade 3. Tab.~\ref{CSATtable} shows results that Grades 1–2 questions were answered with high accuracy, largely because they emphasize fact-checking. In contrast, performance declined on Grade 3 and our constructed dataset, which involve more complex charts and reasoning-intensive questions.

\subsection{Corruption and Occlusion Dataset}
\label{subsec:CorruptandOcclude}

\begin{table}[t]
\centering
\resizebox{\columnwidth}{!}
{
\begin{tabular}{l|l|c|c}

\textbf{\#} & \textbf{Corruption Type} & \textbf{Relevant?} & \textbf{Notes} \\
\hline

1 & Brightness            & Yes     & Affects overall visibility of chart \\
2 & Contrast             & Yes     & Alters legibility, especially in grayscale charts \\
3 & Defocus Blur          & Yes     & Simulates camera/scanner blur \\
4 & Gaussian Blur & Yes   & Softens edges, mimicking poor focus \\
5 & Gaussian Noise        & Yes     & Simulates scan/compression noise \\

6 & Impulse Noise         & Yes     & Strong disruptor of text and lines \\
7 & JPEG Compression     & Yes     & Very common in scanned or saved charts \\
8 & Motion Blur           & Yes     & Mimics shaky camera or scanner \\
9 & Pixelate             & Yes     & Degrades readability at low resolutions \\
10 & Saturate             & Yes     & Distorts color-based encodings \\
11 & Elastic Transform    & Moderate & Artificial but useful for robustness testing \\
12 & Shot Noise            & Moderate & Less realistic for synthetic chart images \\
13 & Spatter              & Moderate & Can simulate water/dirt stains \\
14& Speckle Noise         & Moderate & Common in radar/medical, not typical in charts \\
15 & Zoom Blur             & Moderate & Simulates zoomed or warped chart capture  \\
16 & Glass Blur            & No      & Unnatural distortion for charts \\
17 & Fog                  & No      & Not realistic in document/chart context \\
18 & Frost                & No      & Not applicable to synthetic or scanned charts \\
19 & Snow                 & No      & Not plausible in structured document data \\

\end{tabular}
}
\caption{Overview of corruption types with assessed relevance to charts and graphs}
\label{CorruptionType}
\end{table}


Prior work has applied occlusion to assess object counting\cite{CAPTURE} and has tested corruptions to benchmark VLMs on natural images\cite{BaviBench}. Inspired by these approaches, we extend the idea to chart data, arguing that evaluating VLMs under various noise and occlusion conditions is both suitable and necessary for assessing robustness.

\begin{figure}[ht]
    \begin{center}
        \includegraphics[width=0.95\linewidth]{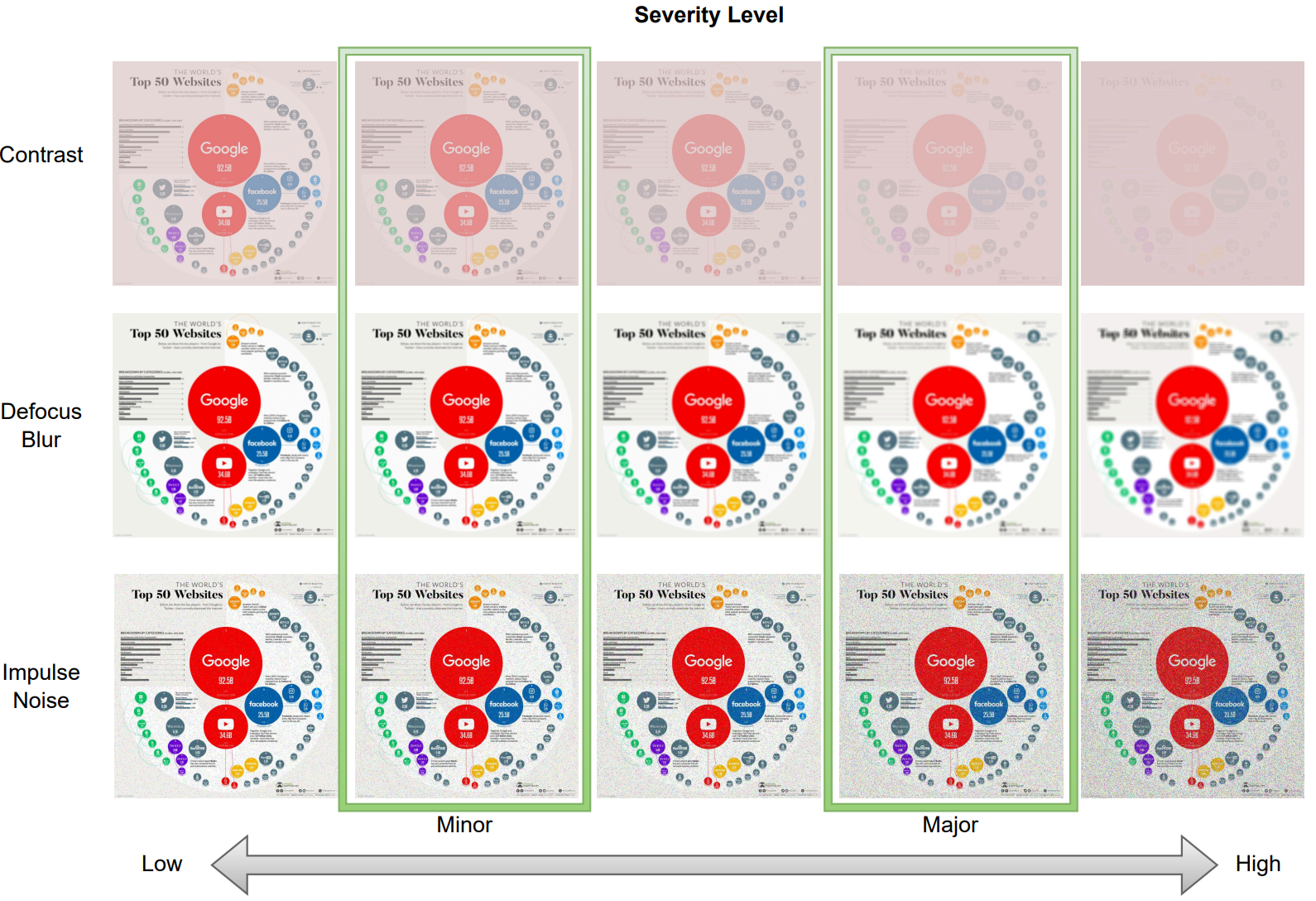}
    \end{center}
\caption{Following the five severity levels defined in~\cite{Corruptions}, we designate level 2 as minor and level 4 as major when generating CHART NOISe dataset}
\label{fig:CorruptionSeverity}
\end{figure}

\begin{figure}[ht]
    \begin{center}
        \includegraphics[width=0.95\linewidth]{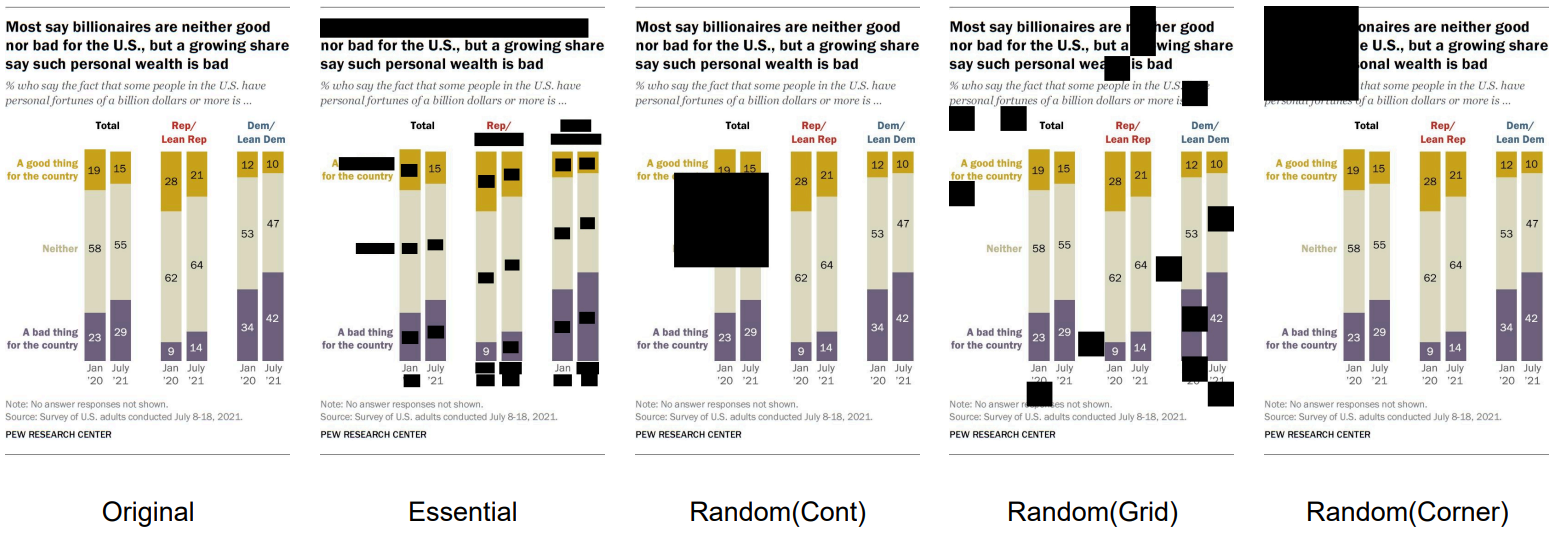}
    \end{center}
\caption{Examples of the four occlusion types in the CHART-NOISe dataset: core, continuous, grid, and corner occlusions}
\label{fig:Occlusion}
\end{figure}

For corruption types, we drew inspiration from Hendrycks and Dietterich's work on benchmarking neural network robustness, which introduced 19 common corruptions and perturbations~\cite{Corruptions}. However, not all are applicable to chart data. As shown in Tab.~\ref{CorruptionType}, We selected 15 corruption types suitable for charts (to be released with the dataset), of which 10 were used in our evaluation. In addition, prior work \cite{Corruptions} defined five severity levels, with examples illustrated in Fig.~\ref{fig:CorruptionSeverity}. For charts, we adopt a simplified view by distinguishing between \textit{major} and \textit{minor} corruption, corresponding to severity levels 2 and 4, respectively.


To identify and occlude essential chart components, we combine attention rollout from a pretrained Vison Transformer (ViT)\cite{VIT} with Optical Character Recognition(OCR) regions extracted using EasyOCR\cite{EASYOCR}, under a strict area budget. We restrict occlusion to a maximum of 7\% of the image, as masking larger portions would obscure too many details and create unrealistic evaluation conditions. Unlike CAM-based methods\cite{zhou2016cam}, which are often noisy and class-specific, attention rollout provides a more stable, model-intrinsic estimate of salient regions. This ensures that high-value textual elements (e.g., axis labels, tick marks, legends) and attention-dense regions are prioritized, while overall context is preserved.

For comparison, we introduce three random occlusion baselines: corner occlusion (peripheral masking), grid occlusion (scattered cell-wise masking), and continuous occlusion (a contiguous block). These provide non-semantic disruptions of comparable area against which to contrast the attention and OCR-guided strategy. An example of chart occlusion is shown in Fig.~\ref{fig:Occlusion}.

\section{Analysis}
\label{sec:analysis}

We evaluated three general-purpose VLMs—ChatGPT-4o, Claude Sonnet 4, and Gemini Pro 2.5. Chart-specific models were excluded, as they are typically smaller, rely on OCR+LLM pipelines, and are not designed for robustness against noise or occlusion. In contrast, the selected large-scale commercial models, though black-box in nature, offer a more rigorous test of susceptibility to such perturbations. This setup parallels robustness evaluation in vision models under data augmentation, allowing us to probe how these systems internally process and interpret charts.

\begin{table}[t]
\centering

\resizebox{\columnwidth}{!}
{
\begin{tabular}{l|ccc|ccc|ccc}

\textbf{Corruption} & \multicolumn{3}{c|}{\textbf{ChatGPT-4o}} & \multicolumn{3}{c|}{\textbf{Claude Sonnet 4}} & \multicolumn{3}{c}{\textbf{Gemini Pro 2.5}} \\
\cline{2-10}
 & Orig & Min & Maj & Orig & Min & Maj & Orig & Min & Maj \\
\hline
Brightness        & 70 & 64 & 63 & 76   &  71   & 70   &  88   & 84   &  81 \\
Contrast          & 70 & 66  & 63 & 76   & 71  &  57(1)  &  88  &  72  &  57(1)  \\
Defocus Blur      & 70 & 53(6) & 54(20) &  76  &  62  &  51(9)  & 88   & 72   &   50(11) \\
Gaussian Blur     & 70 & 61(7) & 45(16) &  76  & 66  &  54(11)  & 88  &  65  &  54(11)   \\
Gaussian Noise    & 70 & 60  & 59 & 76   &  74  & 68   &  88  & 75   &  67   \\
Impulse Noise     & 70 & 61 & 59 &  76  &  70  &  67(2)  &  88  &  70  &  67(2)  \\
JPEG Compression  & 70 & 61 & 61 &   76 &  74  & 75   & 88   &  85  & 85   \\
Motion Blur       & 70 & 59 &  49(6) &  76  & 68(2)   & 45(25)   & 88   &  84   &   77 \\
Pixelate          & 70 & 57 & 55 &  76   &  72  &  72  &  88  & 86 &  79  \\
Saturate          & 70 & 66 & 60 &  76  &  75  &  70  & 88   &  86  & 86   \\

\end{tabular}
}

\caption{Accuracy of ChatGPT-4o, Claude Sonnet 4, and Gemini Pro 2.5 across ten corruption types under original, minor, and major severity levels. Parentheses indicate the number of cases where models requested clarification or flagged quality issues}

\label{Tab3}
\end{table}

\begin{table}[t]
    \centering
    \footnotesize
    \begin{tabular}{l|ccc}
        \hline
        & 
        \textbf{Chatgpt4o} &  \textbf{ClaudeSonnet4} &\textbf{Gemini2.5Pro} \\
        \hline
        Original & 70  & 76  & 88  \\
        Core(targeted) & 60  & 65  &  76  \\
        Random(Cont) & 67   & 68  & 87 \\
        Random(Grid) & 63  & 73 & 83 \\
        Random(Corner) & 67  & 69 & 86 \\
        \hline
    \end{tabular}

\caption{Accuracy of ChatGPT-4o, Claude Sonnet 4, and Gemini Pro 2.5 on the CHART-NOISe dataset under different occlusion strategies: targeted (core), continuous block, grid-based, and corner occlusions, compared to the original condition.}
\label{Tab4}

\end{table}

As described in Section~\ref{subsec:CorruptandOcclude}, we evaluated model robustness and hallucination under 10 corruption types and 4 occlusion strategies. Tab.~\ref{Tab3} presents results across corruption types evaluated at minor and major severity levels, compared with the non-corrupted baseline, while Tab.~\ref{Tab4} summarizes performance under different occlusion settings. As expected, introducing noise and occlusion consistently reduced model accuracy in answering chart-based questions.

\subsection{Research Question 1: What is the evaluation metric?} 
\label{subsec:RQ1}
Since our questions follow the CSAT style, each includes a clearly defined statement that contradicts the chart. We did not impose a token limit on model outputs, nor did we constrain the number of responses. For evaluation, if a model provides multiple answers without isolating the correct one, claims that none of the statements are incorrect, or outputs only an incorrect choice, we count the response as incorrect. This mirrors human evaluation, where identifying the single incorrect statement is required for a correct answer.

\subsection{Research Question 2: What types of explanations do models provide for incorrect answers?}
\label{subsec:RQ2}
Across the evaluated models, we observed several forms of hallucination in their generated explanations, including value fabrication (producing incorrect numerical values), trend misinterpretation (misreading upward or downward patterns), entity confusion (misidentifying chart elements), reasoning hallucination (introducing unsupported logical steps), and table/translation drift (errors during chart-to-table conversion). These errors were further amplified under visual corruptions such as noise or occlusion, leading to more frequent and severe misinterpretations.

\subsection{Research Question 3: How do models perform under visual degradations?}

In addition to clean charts, real-world data often contains visual degradations that hinder model reasoning. 
We investigate two broad categories of degradation: (i) corruption types such as photometric distortions, blur, noise, and compression, and (ii) occlusions that remove essential or non-essential information from the chart. 
Together, these experiments aim to characterize robustness under both quality-based and content-based impairments. This leads to three subquestions:

\vspace{4pt}\noindent\textbf{1) How do models perform under various corruption types?}~~ 

\noindent We categorized corruptions into four groups: Photometric (brightness, contrast, saturate), Blur (defocus, Gaussian, motion), Noise (Gaussian, impulse), and Compression/Resolution Loss (JPEG compression, pixelate). For blur corruptions, models often recognized degraded image quality and requested a clean image (counts in parentheses). In contrast, for other corruption types, accuracy generally declined with increasing severity, but models seldom acknowledged quality issues, frequently providing answers even when the charts were indistinguishable to humans.

\vspace{4pt}\noindent\textbf{2) How do models perform under different occlusions?}~~ 

\noindent We investigated whether occluding essential versus non-essential information affects VLM performance. Simply masking all critical regions would leave models with no meaningful content to interpret, so we limited targeted occlusion (using OCR and attention maps) to approximately 7\% of the chart area. For non-essential information, we assumed peripheral regions contained less critical content and therefore applied corner occlusion. To provide additional baselines, we also introduced grid-based and continuous block occlusions.  

Results show that when essential information is occluded, model accuracy drops noticeably. Surprisingly, however, the models often failed to acknowledge the occlusion, instead generating answers based solely on the remaining unoccluded data. This behavior suggests a tendency toward overconfidence, increasing the likelihood of errors and hallucinations—an issue that warrants caution in practical applications.

\vspace{4pt}\noindent\textbf{3) Do different models exhibit distinct robustness profiles under corruption and occlusion?}~~ 

\noindent Although all models showed vulnerability to corruption and occlusion, their degradation patterns differed. Gemini Pro 2.5 maintained the highest baseline accuracy (88\%) and showed resilience to JPEG compression and corner occlusions, while Claude Sonnet 4 dropped sharply under motion and defocus blur (to 45 and 51 at major severity). ChatGPT-4o was moderately robust but often produced confident answers despite reduced accuracy, highlighting a tendency toward overconfidence.  

Interestingly, these differences reveal that models may have implicit mitigations for certain perturbations yet struggle with others. Even when distortions rendered charts nearly unsolvable to humans, models still attempted to answer, often incorrectly. This suggests that robustness is uneven across systems and that deployment should consider not only average-case performance but also model-specific failure modes under realistic degradations.




\subsection{Research Question 4: Are there ways to mitigate the effects of corruption and occlusion?}  
\label{subsec:RQ4}

For corrupted images, image quality assessment (IQA) metrics such as ARNIQA\cite{agnolucci2024arniqa} have been developed in computer vision. We observed that metric scores consistently decreased from original to minor to major corruptions. With appropriate empirical thresholding, such metrics could be adapted and trained specifically on chart data, serving as a preprocessing filter before passing inputs to a VLM. 

For occlusions, a dedicated occlusion-detection module could be trained not only to identify whether a chart is occluded, but also to segment and localize the missing or obstructed regions, similar to how harmonized objects are localized in forensic tasks by segmentation-based approaches \cite{Shin_2025_WACV}. Charts flagged as occluded could then be filtered, corrected, or augmented with restoration techniques prior to model inference, ensuring that downstream reasoning operates on complete or contextually compensated information. Such front-end filtering approaches offer a potential pathway to improving robustness against both corruption and occlusion.


\section{Discussion}
\label{sec:Discussions}

\subsection {Limitations}
\label{subsec:limitation}
Our study focuses on three large-scale commercial VLMs, leaving out open-source and chart-specific pipelines that may behave differently under corruption and occlusion. Similarly, while we considered ten corruption types and four occlusion strategies, these remain controlled approximations of real-world degradations, which can be more diverse and context-dependent. Finally, the black-box nature of the evaluated models limits our ability to pinpoint whether robustness differences stem from architecture, training data, or inference strategies. These caveats position our findings as a step toward broader robustness evaluation rather than a complete picture, but nonetheless demonstrate the existence of important challenges even if not fully quantifying their scope.  

\subsection{Opportunities and Insights}
\label{subsec:opportunities}

Our results show that VLMs exhibit uneven robustness across corruption types, creating opportunities for targeted training and augmentation. Notably, models often produced confident answers even under severe distortions that humans could not reasonably interpret, underscoring the need for stronger uncertainty calibration. Importantly, \textit{some} sources of distortion are actively identified and mitigated, implying that a comprehensive solution could be implementable. Practical improvements may come from front-end modules such as quality filters or occlusion detectors, while the exam-style setting of CHART NOISe frames chart reasoning as a joint challenge of vision, reasoning, and reliability.
\section{Conclusion}
\label{sec:conclusion}
In this work, we introduced a new dataset and evaluation framework for probing the robustness of vision–language models (VLMs) in chart understanding. By integrating chart corruptions, occlusions, and CSAT-style multiple-choice questions that exploit prompt reverse inconsistency, our benchmark highlights vulnerabilities that are often overlooked in existing datasets. Experiments with state-of-the-art VLMs(ChatGPT, Gemini and Claude Sonnet) show that, while models perform well on clean and fact-oriented questions, their accuracy drops significantly under visual degradation and reasoning-intensive settings, often accompanied by errors such as value fabrication, trend misinterpretation, and entity confusion.

Our findings underscore the need for robust evaluation protocols that go beyond surface-level chart parsing to capture both visual resilience and reasoning consistency. As a step forward, we suggest incorporating preprocessing filters, occlusion detection modules, and quality-aware pipelines as practical mitigation strategies. We intend our dataset to serve as a foundation for developing VLMs that are not only capable of understanding charts but also reliable under real-world conditions where imperfections and ambiguities are inevitable.

\section{Acknowledgement}
\label{sec:acknowledgment}

This work was supported by the U.S. DOE Office of Science, Office of Advanced Scientific Computing Research, under award 76125: ``AMAIS - Advanced Memory to support Artificial Intelligence for Science''. The Pacific Northwest National Laboratory is operated by Battelle for the U.S. Department of Energy under contract DE-AC05-76RL01830, and in part by NSF Awards 2243979


\bibliographystyle{IEEEbib}
\bibliography{strings,refs}

\begin{thebibliography}{10}

\bibitem{ChartQARL}
Yuyang Ji and Haohan Wang,
\newblock ``Socratic chart: Cooperating multiple agents for robust svg chart understanding,''
\newblock {\em arXiv preprint arXiv:2504.09764}, 2025.

\bibitem{wang2024charxiv}
Zirui Wang, Mengzhou Xia, Luxi He, Howard Chen, Yitao Liu, Richard Zhu, Kaiqu Liang, Xindi Wu, Haotian Liu, Sadhika Malladi, et~al.,
\newblock ``Charxiv: Charting gaps in realistic chart understanding in multimodal llms,''
\newblock {\em Advances in Neural Information Processing Systems}, vol. 37, pp. 113569--113697, 2024.

\bibitem{kice}
{Korea Institute for Curriculum and Evaluation (KICE)},
\newblock ``Official website of the korea institute for curriculum and evaluation,'' \url{http://www.kice.re.kr}.

\bibitem{Reverseprompt}
Jihyun~Janice Ahn and Wenpeng Yin,
\newblock ``Prompt-reverse inconsistency: Llm self-inconsistency beyond generative randomness and prompt paraphrasing,''
\newblock {\em arXiv preprint arXiv:2504.01282}, 2025.

\bibitem{gemini2.5pro}
Google DeepMind,
\newblock ``Gemini 2.5 pro,'' \url{https://deepmind.google}, 2025,
\newblock Large language model.

\bibitem{chatgpt4o}
OpenAI,
\newblock ``Chatgpt-4o,'' \url{https://chat.openai.com}, 2025,
\newblock Large language model.

\bibitem{claude-sonnet4}
Anthropic,
\newblock ``Claude sonnet 4.0,'' \url{https://claude.ai}, 2025,
\newblock Large language model.

\bibitem{deplot}
Fangyu Liu, Julian Eisenschlos, Francesco Piccinno, Syrine Krichene, Chenxi Pang, Kenton Lee, Mandar Joshi, Wenhu Chen, Nigel Collier, and Yasemin Altun,
\newblock ``Deplot: One-shot visual language reasoning by plot-to-table translation,''
\newblock in {\em Findings of the Association for Computational Linguistics: ACL 2023}, 2023, pp. 10381--10399.

\bibitem{unichart}
Ahmed Masry, Parsa Kavehzadeh, Xuan~Long Do, Enamul Hoque, and Shafiq Joty,
\newblock ``Unichart: A universal vision-language pretrained model for chart comprehension and reasoning,''
\newblock {\em arXiv preprint arXiv:2305.14761}, 2023.

\bibitem{chartassisstant}
Fanqing Meng, Wenqi Shao, Quanfeng Lu, Peng Gao, Kaipeng Zhang, Yu~Qiao, and Ping Luo,
\newblock ``Chartassisstant: A universal chart multimodal language model via chart-to-table pre-training and multitask instruction tuning,''
\newblock {\em CoRR}, 2024.

\bibitem{tinychart}
Liang Zhang, Anwen Hu, Haiyang Xu, Ming Yan, Yichen Xu, Qin Jin, Ji~Zhang, and Fei Huang,
\newblock ``Tinychart: Efficient chart understanding with visual token merging and program-of-thoughts learning,''
\newblock {\em CoRR}, 2024.

\bibitem{survey1}
Zechen Bai, Pichao Wang, Tianjun Xiao, Tong He, Zongbo Han, Zheng Zhang, and Mike~Zheng Shou,
\newblock ``Hallucination of multimodal large language models: A survey,'' 2025.

\bibitem{survey2}
Hanchao Liu, Wenyuan Xue, Yifei Chen, Dapeng Chen, Xiutian Zhao, Ke~Wang, Liping Hou, Rongjun Li, and Wei Peng,
\newblock ``A survey on hallucination in large vision-language models,'' 2024.

\bibitem{park2025second}
Woohyeon Park, Woojin Kim, Jaeik Kim, and Jaeyoung Do,
\newblock ``Second: Mitigating perceptual hallucination in vision-language models via selective and contrastive decoding,''
\newblock {\em arXiv preprint arXiv:2506.08391}, 2025.

\bibitem{zhao2024mitigating}
Linxi Zhao, Yihe Deng, Weitong Zhang, and Quanquan Gu,
\newblock ``Mitigating object hallucination in large vision-language models via image-grounded guidance,''
\newblock {\em arXiv preprint arXiv:2402.08680}, 2024.

\bibitem{liu2024reducing}
Sheng Liu, Haotian Ye, Lei Xing, and James Zou,
\newblock ``Reducing hallucinations in vision-language models via latent space steering,''
\newblock {\em arXiv preprint arXiv:2410.15778}, 2024.

\bibitem{BaviBench}
Hao Zhang, Wenqi Shao, Hong Liu, Yongqiang Ma, Ping Luo, Yu~Qiao, Nanning Zheng, and Kaipeng Zhang,
\newblock ``B-avibench: Toward evaluating the robustness of large vision-language model on black-box adversarial visual-instructions,''
\newblock {\em Trans. Info. For. Sec.}, vol. 20, pp. 1434–1446, Jan. 2025.

\bibitem{methani2020plotqa}
Nitesh Methani, Pritha Ganguly, Mitesh~M Khapra, and Pratyush Kumar,
\newblock ``Plotqa: Reasoning over scientific plots,''
\newblock in {\em Proceedings of the ieee/cvf winter conference on applications of computer vision}, 2020, pp. 1527--1536.

\bibitem{ChartQAPro}
Ahmed Masry, Mohammed~Saidul Islam, Mahir Ahmed, Aayush Bajaj, Firoz Kabir, Aaryaman Kartha, Md~Tahmid~Rahman Laskar, Mizanur Rahman, Shadikur Rahman, Mehrad Shahmohammadi, Megh Thakkar, Md~Rizwan Parvez, Enamul Hoque, and Shafiq Joty,
\newblock ``Chartqapro: A more diverse and challenging benchmark for chart question answering,'' 2025.

\bibitem{masry2022chartqa}
Ahmed Masry, Do~Xuan Long, Jia~Qing Tan, Shafiq Joty, and Enamul Hoque,
\newblock ``Chartqa: A benchmark for question answering about charts with visual and logical reasoning,''
\newblock {\em arXiv preprint arXiv:2203.10244}, 2022.

\bibitem{CAPTURE}
Atin Pothiraj, Elias Stengel-Eskin, Jaemin Cho, and Mohit Bansal,
\newblock ``Capture: Evaluating spatial reasoning in vision language models via occluded object counting,'' 2025.

\bibitem{Corruptions}
Dan Hendrycks and Thomas Dietterich,
\newblock ``Benchmarking neural network robustness to common corruptions and perturbations,''
\newblock {\em Proceedings of the International Conference on Learning Representations}, 2019.

\bibitem{VIT}
Alexey Dosovitskiy, Lucas Beyer, Alexander Kolesnikov, Dirk Weissenborn, Xiaohua Zhai, Thomas Unterthiner, Mostafa Dehghani, Matthias Minderer, G~Heigold, S~Gelly, et~al.,
\newblock ``An image is worth 16x16 words: Transformers for image recognition at scale,''
\newblock in {\em International Conference on Learning Representations}, 2020.

\bibitem{EASYOCR}
Jaided AI,
\newblock ``Easyocr,'' \url{https://github.com/JaidedAI/EasyOCR}, 2020.

\bibitem{zhou2016cam}
Bolei Zhou, Aditya Khosla, Agata Lapedriza, Aude Oliva, and Antonio Torralba,
\newblock ``Learning deep features for discriminative localization,''
\newblock in {\em Proceedings of the IEEE Conference on Computer Vision and Pattern Recognition (CVPR)}, 2016, pp. 2921--2929.

\bibitem{agnolucci2024arniqa}
Lorenzo Agnolucci, Leonardo Galteri, Marco Bertini, and Alberto Del~Bimbo,
\newblock ``Arniqa: Learning distortion manifold for image quality assessment,''
\newblock in {\em Proceedings of the IEEE/CVF Winter Conference on Applications of Computer Vision}, 2024, pp. 189--198.

\bibitem{Shin_2025_WACV}
Philip~Wootaek Shin, Jack Sampson, Vijaykrishnan Narayanan, Andres Marquez, and Mahantesh Halappanavar,
\newblock ``Disharmony: Forensics using reverse lighting harmonization,''
\newblock in {\em Proceedings of the Winter Conference on Applications of Computer Vision (WACV) Workshops}, February 2025, pp. 756--765.

\end{thebibliography}

\clearpage
\setcounter{page}{1}
\maketitlesupplementary
\appendix

\section{Corruption Types for Charts} 

\begin{figure}[h]
    \begin{center}
        \includegraphics[width=0.87\linewidth]{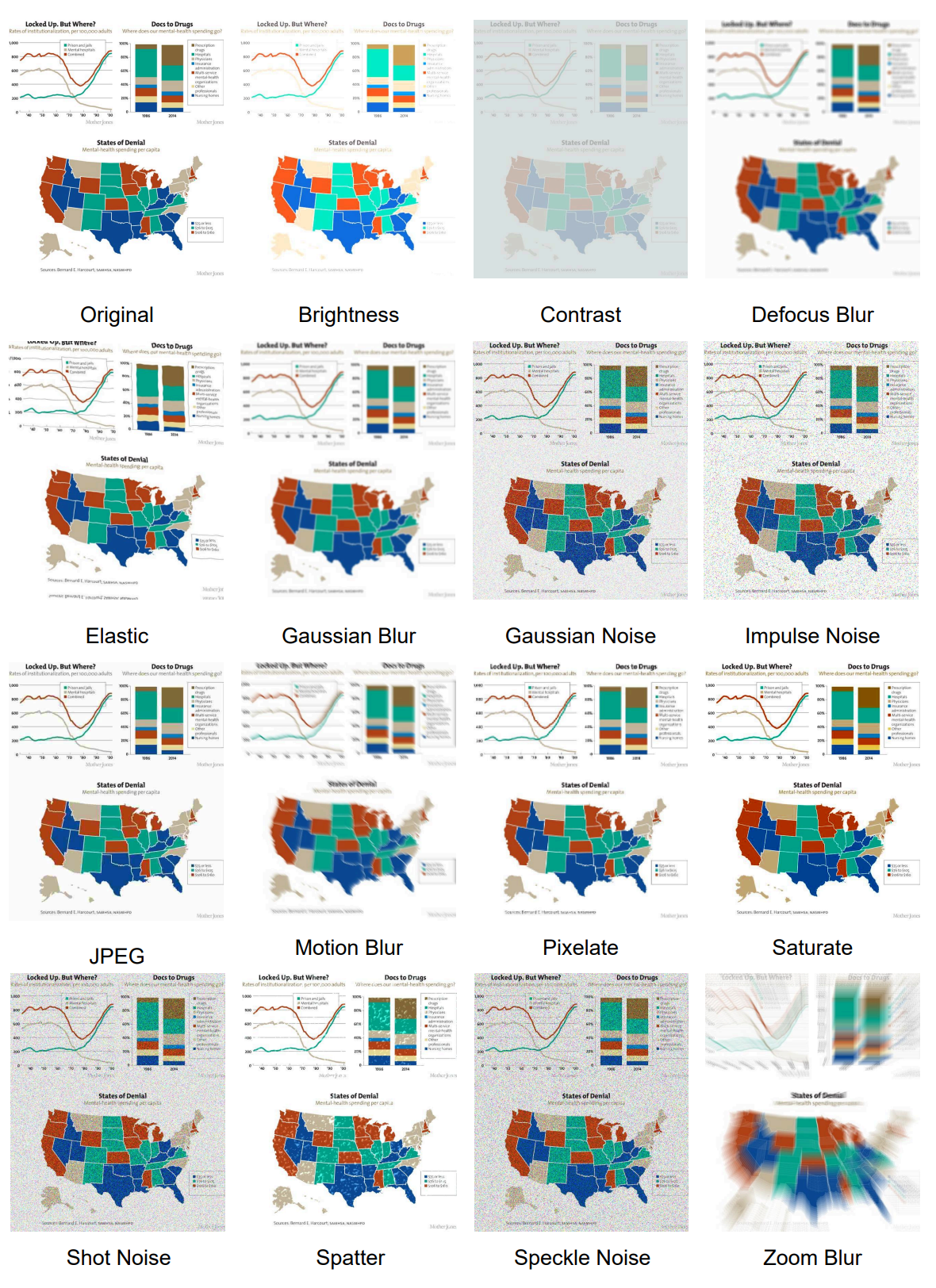}
    \end{center}
\caption{Illustration of 15 corruption types relevant to charts, shown at minor severity level.}
\label{fig:CorruptionType}
\end{figure}

As mentioned in Sec.~\ref{subsec:CorruptandOcclude}, we selected 15 corruption types suitable for charts, with the rationale summarized in Tab.~\ref{CorruptionType}, of which 10 were used in our evaluation. To provide a clearer visualization, Fig.~\ref{fig:CorruptionType} illustrates all 15 corruption types applied at minor severity. While our evaluation focused on 10 corruptions across different VLMs, the dataset includes examples of all 15 types for broader analysis. The occlusion types are all shown in Fig.~\ref{fig:Occlusion}; therefore, we do not provide additional figures for separate examples.

\section{Model Evaluation Setup}

For evaluation, we tested three proprietary VLMs: ChatGPT-4o \cite{chatgpt4o}, Gemini 2.5 Pro \cite{gemini2.5pro}, and Claude Sonnet 4 \cite{claude-sonnet4}. While open-source VLMs could have been included, we prioritized these commercial systems, as they are continually refined through user feedback and expected to be more robust in practical deployments. Chart-specific models discussed in Sec.~\ref{subsec:Chart} were excluded after preliminary trials, since under degraded conditions they frequently produced unreliable or incoherent answers.In particular, when presented with corrupted or partially occluded figures, these models often failed to extract basic chart elements, misaligned values with axes, or generated outputs that contradicted the visual evidence. Such brittleness not only undermined their comparability with general-purpose VLMs but also made their results inconsistent across repeated runs, highlighting a lack of stability in real-world, imperfect scenarios.

Unlike many prior studies, we did not restrict output length (Sec.~\ref{subsec:RQ1}). We found that enforcing strict token limits often led to incorrect outputs. For instance, Gemini 2.5 Pro sometimes initially chose the wrong option but then engaged in a self-correction process, revisiting its reasoning before arriving at the correct answer. When output length was restricted, this reflective process was truncated, and the model more frequently produced faulty responses.

Finally, although our multiple-choice setting contained a single incorrect statement among five options, we did not explicitly constrain models to output exactly one choice. This design allowed us to observe how models behaved when given flexibility: Gemini tended to identify the best single choice even without explicit instruction, while ChatGPT-4o and Claude Sonnet 4 occasionally selected multiple options or provided less consistent justifications. These behavioral differences highlight variation in how models internally process ambiguity and reason about chart-based tasks.

\begin{table}[h]
\centering
\resizebox{\columnwidth}{!}{
\footnotesize
\begin{tabular}{l|c|c|c}
\hline
\textbf{Category} & \textbf{Minor } & \textbf{Major } & \textbf{Change (Min $\rightarrow$ Maj)} \\
\hline
Brightness      & 0.275 & 0.236 & $-0.039$ \\
Contrast        & 0.263 & 0.190 & $-0.073$ \\
Defocus Blur    & 0.175 & 0.084 & $-0.091$ \\
Gaussian Blur   & 0.198 & 0.133 & $-0.065$ \\
Gaussian Noise  & 0.313 & 0.181 & $-0.132$ \\
Impulse Noise   & 0.275 & 0.166 & $-0.109$ \\
JPEG Compression& 0.310 & 0.198 & $-0.112$ \\
Motion Blur     & 0.246 & 0.196 & $-0.050$ \\
Pixelate        & 0.141 & -0.026& $-0.167$ \\
Saturate        & 0.301 & 0.233 & $-0.068$ \\

\hline
\multicolumn{4}{c}{\textbf{Original: 0.327}} \\
\hline

\end{tabular}
}
\caption{Average ARNIQA\cite{agnolucci2024arniqa} Scores under Minor and Major corruptions on chart data}
\label{QualityMetric}
\end{table}

\begin{figure}[ht]
    \begin{center}
        \includegraphics[width=0.6\linewidth]{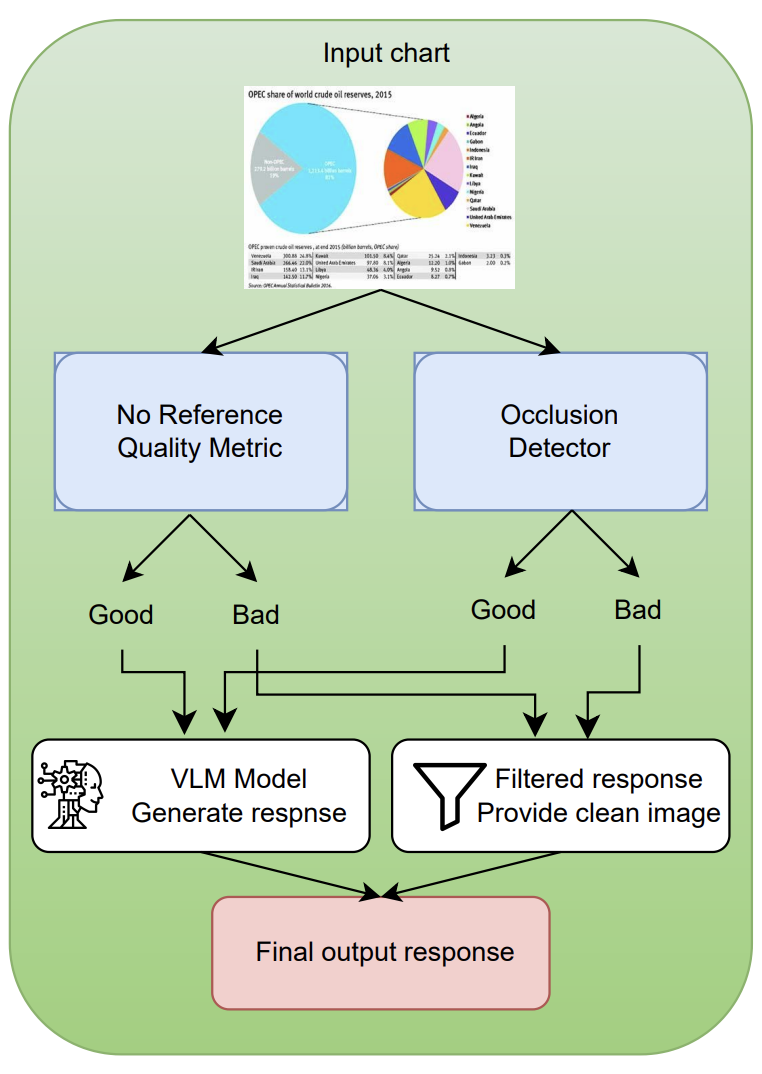}
    \end{center}
\caption{Filtering method applied to the chart data}
\label{fig:Mitigation}
\end{figure}

\section{Mitigation of Corruptions and Occlusions}

Figure~\ref{fig:Mitigation} summarizes our front-end mitigation pipeline. Prior to VLM inference, chart images are screened to detect (i) corruption and (ii) occlusion, with the goal of intercepting unreliable inputs that would otherwise elicit spurious or hallucinated answers. Although this introduces modest preprocessing overhead, it ultimately saves compute and reduces error rates by avoiding full VLM decoding on low-quality inputs.

Image quality assessment (IQA) metrics can serve as a chart-aware gate. Table~\ref{QualityMetric} reports ARNIQA~\cite{agnolucci2024arniqa} scores under minor and major corruptions, where we observe a consistent monotonic drop from original \(\rightarrow\) minor \(\rightarrow\) major. While a universal threshold is unlikely, empirical calibration on development data can set practical cutoffs for flagging or rejecting corrupted inputs, and chart-specific fine-tuning of IQA models can further improve sensitivity.

For occlusions, we adopt a segmentation-based approach inspired by Shin et al.~\cite{Shin_2025_WACV}. In our experiments, this framework successfully localized occluded regions on charts; the resulting masks allow us to quantify the occluded fraction and use it as a filter (e.g., reject or route to restoration when the fraction exceeds a preset threshold). This ensures downstream reasoning operates on complete—or explicitly compensated—visual evidence.

Some proprietary VLMs occasionally self-report low image quality, but we found this behavior to be largely confined to blur; it rarely triggers for noise, compression, or occlusion. External screening is therefore advisable. More broadly, models often produced confident answers under severe degradations that humans could not reasonably interpret, underscoring the need for stronger uncertainty calibration alongside the proposed front-end filters.

\section{Hallucination in Chart Understanding}

As detailed in Sec.~\ref{subsec:RQ2}, we observed several forms of hallucination in model explanations, including value fabrication, trend misinterpretation, entity confusion, reasoning hallucination, and table/translation drift, with these errors further amplified under corruptions and occlusions. Although we can provide representative examples, we leave a more systematic study of chart hallucination to future work. At present, hallucination in this domain is not clearly defined, and establishing a rigorous framework for categorization and measurement remains an open challenge. We view this as an important research problem and a promising direction for continued exploration.

\end{document}